\documentclass{article}

\usepackage{PRIMEarxiv}

\usepackage[utf8]{inputenc} % allow utf-8 input
\usepackage[T1]{fontenc}    % use 8-bit T1 fonts
\usepackage{hyperref}       % hyperlinks
\usepackage{url}            % simple URL typesetting
\usepackage{booktabs}       % professional-quality tables
\usepackage{amsfonts}       % blackboard math symbols
\usepackage{nicefrac}       % compact symbols for 1/2, etc.
\usepackage{microtype}      % microtypography
\usepackage{lipsum}
\usepackage{fancyhdr}       % header
\usepackage{graphicx}       % graphics
\usepackage{amsmath}
\usepackage{xcolor}
\usepackage{mathtools}
\usepackage{textcomp}
\usepackage{caption}
\usepackage{subcaption}
\usepackage{comment}

\title{Promoting Generalized Cross-lingual Question Answering in Few-resource Scenarios via Self-knowledge Distillation}
\author{
  Casimiro Pio Carrino, Carlos Escolano, José A. R. Fonollosa \\
  Universitat Politècnica De Catalunya \\
  \texttt{\{casimiro.pio.carrino,carlos.escolano,jose.fonollosa\}@upc.edu} \\
  }

\begin{document}
\maketitle
\begin{abstract}
Despite substantial progress in multilingual extractive Question Answering (QA), models with high and uniformly distributed performance across languages remain challenging, especially for languages with limited resources. We study cross-lingual transfer mainly focusing on the Generalized Cross-Lingual Transfer (G-XLT) task, where the question language differs from the context language — a challenge that has received limited attention thus far.
Our approach seeks to enhance cross-lingual QA transfer using a high-performing multilingual model trained on a large-scale dataset, complemented by a few thousand aligned QA examples across languages. We build our techniques upon the analysis of the cross-lingual transfer capabilities of a pre-trained multilingual BERT model fine-tuned on English-language SQuAD-v1.1. Our proposed strategy combines cross-lingual sampling and advanced self-distillation training in generations to tackle the previous challenge. Notably, we introduce the novel {\em{mAP@k coefficients}} to fine-tune self-knowledge distillation loss, dynamically regulating the teacher's model knowledge to perform a balanced and effective knowledge transfer. We extensively evaluate our approach using various QA datasets, including MLQA, XQuAD, and TyDiQA-goldp, to assess XLT and G-XLT capabilities in extractive QA. Results reveal that our self-knowledge distillation approach outperforms standard cross-entropy fine-tuning by a significant margin. Importantly, when compared to a strong baseline that leverages a sizeable volume of machine-translated data, our approach shows competitive results despite the considerable challenge of operating within resource-constrained settings, even in zero-shot scenarios. Beyond performance improvements, we offer valuable insights through comprehensive analyses and an ablation study, further substantiating the benefits and constraints of our approach. In essence, we propose a practical solution to improve cross-lingual QA transfer by leveraging a few data resources in an efficient way. 
\end{abstract}

% keywords can be removed
% \keywords{cross-lingual question answering \and self-knowledge distillation}

% a method that is especially valuable for languages with limited data resources where small-scale annotation efforts are feasible.
\section{Introduction} \label{sec:intro}
Significant advancements have been made in the realm of cross-lingual Question Answering QA in recent years, attributed to the emergence of powerful cross-lingual representations acquired through multilingual Pre-trained Language Models (PLMs) \cite{devlin_bert_2019,conneau-etal-2020-unsupervised,conneau_cross-lingual_2019,xue-etal-2021-mt5,huang_unicoder_2019}. Transfer learning techniques have enriched Cross-lingual Transfer (XLT) capabilities by seamlessly enabling PLMs to operate across languages. The realm of enhancing Cross-Lingual Transfer (XLT) within the domain of Question Answering (QA) has witnessed a diverse array of methodologies. These methods encompass various strategies, including zero-shot and few-shot transfer techniques \cite{hsu_zero-shot_2019,nooralahzadeh_zero-shot_2020,liu_cross-lingual_2020,lewis_mlqa_2020,lauscher_zero_2020}, the utilization of machine-translation and data augmentation strategies \cite{cui_cross-lingual_2019,lee_cross-lingual_2019,liu_cross-lingual_2020,singh_xlda_2019,riabi-etal-2021-synthetic}, as well as more sophisticated approaches such as meta-learning \cite{nooralahzadeh_zero-shot_2020}, leveraging knowledge graphs \cite{duan_bridging_2021}, employing contrastive learning \cite{chen_good_2022}, and incorporating adversarial training \cite{bornea_multilingual_2021}, and defining auxiliary tasks for PLMs \cite{yuan_enhancing_2020}.  Concurrently, substantial efforts have been invested in the development of comprehensive and rigorous benchmarks that aim to assess cross-lingual transferability in the context of the QA task \cite{clark_tydi_2020,lewis_mlqa_2020,artetxe_cross-lingual_2020,asai_xor_2021,longpre_mkqa_2021,liu-etal-2019-xqa}.

However, the pursuit of attaining cross-lingual QA performance characterized by high and uniformly distributed proficiency across languages persists as a formidable challenge. This challenge is especially pronounced in languages constrained by limited linguistic resources. Of particular significance is the aspiration for achieving generalized cross-lingual transfer (G-XLT), where the language of the posed question is different from the language of the answer. This specific avenue of inquiry assumes importance in addressing the challenges posed by information scarcity and linguistic asymmetry in languages with limited resources \cite{asai_xor_2021}, and gains relevance in situations where language mismatch could hinder effective information extraction \cite{Roy2020}. However, it is worth noting that this area of investigation remains largely unexplored, with only a handful of studies dedicated to its exploration.

This study focuses on the enhancement of cross-lingual abilities for the extractive QA task, with particular emphasis on advancing the G-XLT capabilities. The methodology employed involves the transfer of QA knowledge extracted from a proficient multilingual QA model, which has undergone fine-tuning using a large-scale QA dataset in a high-resource language. Remarkably, we delve into attaining effective knowledge transfer by harnessing as few as a thousand QA examples aligned across languages to aid and promote the transfer. We commence by scrutinising the zero-shot knowledge of an mBERT model fine-tuned on the SQuAD-v1.1 dataset in English, which provided essential insights for the design of our strategy. Then, we tackle the challenge by proposing a customized cross-lingual QA fine-tuning strategy that involves cross-lingual sampling and self-distillation training, a special case of knowledge distillation where the teacher model becomes the student model itself. Importantly, we introduce the {\em{mAP@k coefficients}} to modulate the self-knowledge distillation loss. These coefficients dynamically regulate the influence of the teacher's cross-lingual knowledge during the fine-tuning process, thereby facilitating a balanced knowledge transfer. Ultimately, we conduct a comprehensive assessment of our methodology, employing a diverse array of QA benchmarks such as MLQA, XQuAD, and TyDiQA-golp. Our objective is to scrutinize the extent of XLT and G-XLT capabilities that our approach demonstrates in the context of extractive QA. Additionally, our work provides valuable insights, supported by thorough analyses and an ablation study, emphasising the strengths and limitations of our approach.

% 5) State the contributions
In summary, our study's key contributions are as follows:

\begin{itemize}
\item We introduce effective self-knowledge distillation techniques tailored for cross-lingual fine-tuning, utilizing aligned multilingual QA data and cross-lingual sampling to bolster knowledge transfer between languages.

\item We propose the mAP@k loss coefficients to better handle wrong teacher predictions, making the cross-lingual transfer more robust and resulting in enhanced XLT and G-XLT performances, including zero-shot scenarios.

\item We perform a comprehensive analysis and ablation study, unveiling the influence of distinct components and design choices to elucidate the underlying mechanisms behind our approach. 
\end{itemize}

Therefore, our investigation lays the foundation for enhancing cross-lingual QA transfer efficacy in data-scarce scenarios. Besides, we believe the introduction of the mAP@k coefficients may be of potential interest in different knowledge distillation settings beyond QA applications.

\section{Related Works}
% Summarizes and critically evaluates relevant previous research relevant to this work
% Divide it into sections if necessary
In this section, we describe most similar studies utilising fine-tuning techniques for extractive QA to enhance G-XLT and XLT performances on common benchmarks. In \cite{liu_cross-lingual_2020} the authors present the concept of Language Branch Machine Reading Comprehension (LBMRC), which employs language-specific branches, to sample data that pair passages in a single language to questions in various other languages using machine translation, along with a novel multilingual multi-teacher distillation framework. This approach effectively improves cross-lingual performance and enhances robustness against noise in low-resource languages such as Arabic, Hindi, and Vietnamese. The proposed methodology achieves remarkable outcomes on the XQuAD \cite{artetxe_cross-lingual_2020} and MLQA \cite{lewis_mlqa_2020} benchmarks for cross-lingual QA, both in translation and zero-shot scenarios, underscoring its efficacy in addressing the challenges associated with QA tasks. In \cite{bornea_multilingual_2021}, the authors enhance cross-lingual QA transfer by augmenting training data by 14 times through machine translation, language adversarial training, and a Language Arbitration Framework (LAF). They empirically validated on MLQA \cite{lewis_mlqa_2020} and TyDiQA \cite{clark_tydi_2020} datasets, demonstrating significant improvements over the zero-shot baseline in \cite{lewis_mlqa_2020}, highlighting its limitations. In contrast to these investigations, our approach is distinct in that it relies solely on a few thousand cross-lingual QA examples, devoid of a substantial machine-translated dataset, thus posing extra challenges. Additionally, we adopt a novel self-distillation approach that leverages customized mAP@k loss coefficients to further enhance the efficiency of the transfer learning procedure.
% cite methods that employ Multilingual fine-tuning (using multilingual data)
%1) Liu et al., “Cross-Lingual Machine Reading Comprehension with Language Branch Knowledge Distillation.”
% 2) Bornea et al., “Multilingual Transfer Learning for QA Using Translation as Data Augmentation.”
% 3) Zheng et al., “Consistency Regularization for Cross-Lingual Fine-Tuning.”
% 4) Chen et al., “From Good to Best.”
\section{Solving the Extractive QA task} 
\label{sec:extactive-qa-task}
The extractive question-answering (QA) task involves a question $q$, a context $c$, and an answer $a$ that corresponds to a span within the context $c$. To solve the task, the standard method in \cite{devlin_bert_2019} employs a classification layer on top of a transformer-based pre-trained encoder. First, the input question and the context are concatenated into a single sequence and encoded through contextualized embeddings of a given dimension ${T_k} \in R^h$. Then, for each context token $i$, we compute the probabilities $p_{k}^{start}$ and $ p_{k}^{end}$ of being the start and end token of the answer span $a$, respectively, with a softmax over all the tokens ${T_m}$ in the context:
    
\begin{equation}
\begin{split}
    p_{k}^{start} = softmax(e^{S  T_k}; t=1) = \frac{e^{S  T_k/t}}{\sum_m e^{S  T_m/t}} \\
    p_{k}^{end} = softmax(e^{E  T_i}; t=1) =  \frac{e^{E  T_k/t}}{\sum_m e^{E  T_m/t}}
    \label{prob_start}
\end{split}
\end{equation}
Where $T_k$ is the contextualized embedding of the token {\em{k}}, $S \in \mathcal{R}^h$ and $E \in \mathcal{R}^h$ are the start and end vectors representing the trainable parameters of the classification layer, and $t$ is the temperature of the softmax function. Then, for each example, the total loss to optimize is the sum over the context tokens $i \in {1, N}$ of the cross-entropy (CE) between the ground-truth labels $\{a_i^l\}$ and the model probabilities $\{p_i^l\}$, as follows:

\begin{equation} 
    L_{ce} = \sum_{l=start}^{end} \sum_{i=1}^{N} log(p_i^l) = \sum_{l=start}^{end} \sum_{i=1}^{N} CE(a_i^l, p_i^l)
\end{equation}
\label{eq:ce-loss}

Following training, the model identifies the answer by selecting the span defined by the start and end tokens with the highest probabilities. It ensures that potential spans where the end tokens are predicted before the start tokens are excluded.

\section{Proposed Approach}
In this section, we introduce our self-distillation approach to achieve cross-lingual QA performance by transferring the QA knowledge from a high-resource language, such as English, to multiple low-resource target languages. We initiate our study by analyzing the zero-shot cross-lingual performance of the extractive QA task to assess its limitations and gain valuable insights that set the base of our work. Subsequently, we introduce a tailored cross-lingual sampling method that leverages parallel QA datasets aligned across multiple languages, to foster the development of robust G-XLT capabilities. Importantly, we propose a self-knowledge distillation fine-tuning using a pre-trained model by endowing the loss function with the mAP@k loss coefficients to balance multi-loss contributions.

\subsection{A close look at Zero-shot Cross-lingual Transfer}
\label{sec:zero-shot}
We choose the widely employed mBERT \cite{DBLP:conf/naacl/DevlinCLT19} model as our baseline and utilize the SQuAD v1.1 dataset \cite{rajpurkar-etal-2016-squad}, a well-established high-resource extractive QA dataset in English. Specifically, we conduct fine-tuning of the mBERT model using the SQuAD v1.1 dataset using the standard approach as described in Section \ref{sec:extactive-qa-task}. We performed fine-tuning for 3 epochs, with a learning rate of 3e-5 and a batch size of 24. The rest of the hyperparameters are set to their default values \footnote{This standard configuration of the hyperparameters aims to provide a standard reference point for comparison and ensures consistency with previous research.} as implemented in the popular Hugging Face's Transformers library \cite{wolf2020huggingfaces}. We refer to the resulting model as the mBERT-qa-en model. 

\begin{figure}[!h]
    \centering
    \begin{subfigure}[b]{0.49\textwidth}
    \includegraphics[scale=0.50]{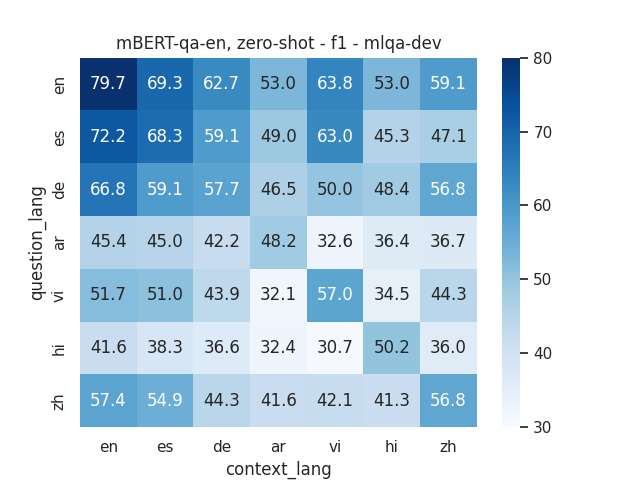}
    \end{subfigure}
    \begin{subfigure}[b]{0.49\textwidth}
    \includegraphics[scale=0.50]{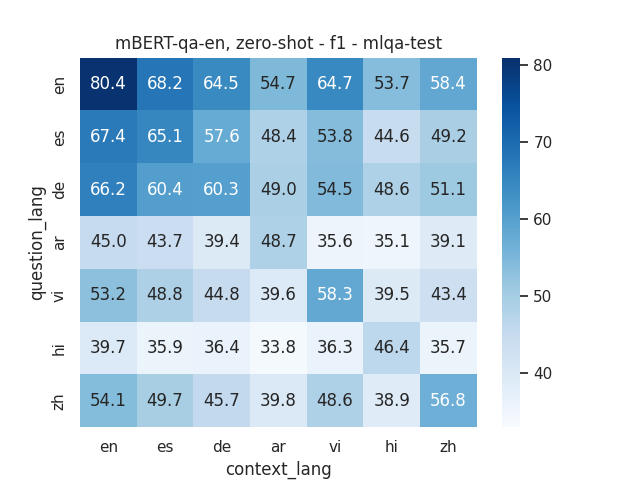}
    \end{subfigure}
    \caption{Zero-shot F1 performance of the mBERT-qa-en for the G-XLT evaluation on the MLQA dev and test datasets.}
    \label{fig:mbert-qa-en-zero-shot-base}
\end{figure}

\paragraph{Measuring the G-XLT performance}
As stated in the Introduction, our primary objective is to achieve cross-lingual extractive QA performance that surpasses the limitations of the same-language setting, particularly in scenarios where questions and the corresponding contextual information are expressed in different languages. To assess this capability, we commence by measuring the generalized cross-lingual transfer (G-XLT) performance of the mBERT-qa-en model on the MLQA benchmark \cite{lewis_mlqa_2020}. Results on the dev and test splits, as illustrated in Figure \ref{fig:mbert-qa-en-zero-shot-base}, reveal a high degree of transferability for languages closely related to English, such as Spanish and German, with a maximum F1 score of 71.0. However, when confronted with less similar languages, particularly in cases where the language of the question differs from the language of the context, the performance diminishes significantly, reaching a minimum F1 score as low as 29.7. In Appendix \ref{app:em-scores}, we show similar patterns in the corresponding Exact Match (EM).

\paragraph{Hidden Knowledge in the Top-k Answer Predictions}
To further measure the extent of cross-lingual knowledge embedded within the mBERT-qa-en model, we conduct an analysis of the prediction quality derived from the top $k$ answer predictions, partially building upon the methodology in \cite{chen_good_2022}. Specifically, we quantify the number of predictions among the top 10 ranked answers that are correct, indicating the presence of hidden knowledge that may be advantageous for subsequent cross-lingual fine-tuning. Consequently, we calculate the distribution of these predictions with their respective ranks within the top 10 positions. Moreover, we sort the precitions based on the language of the corresponding questions to gain insights into their distribution across different languages. As shown in Table \ref{tab:top-k-answers}, our findings highlight the presence of a significant number of correct predictions hidden beyond the top-1 position exhibiting a decreasing trend towards the tenth position.

\begin{table}[!ht]
    \centering
    \small
    \begin{tabular}{c|lllllll}
    \hline
        &\multicolumn{7}{c}{\textbf{No. correct predictions}} \\ \hline
        top $k$ & en & es & de & ar & vi & hi & zh \\ 
        1 & 1773 & 720 & 684 & 492 & 525 & 442 & 592 \\ 
        2 & 460 & 215 & 195 & 136 & 175 & 144 & 170 \\ 
        3 & 272 & 99 & 106 & 82 & 98 & 89 & 94 \\ 
        4 & 188 & 86 & 89 & 57 & 55 & 72 & 77 \\ 
        5 & 135 & 60 & 43 & 68 & 48 & 47 & 61 \\ 
        6 & 92 & 34 & 54 & 28 & 42 & 35 & 44 \\ 
        7 & 68 & 42 & 32 & 45 & 46 & 31 & 30 \\ 
        8 & 68 & 28 & 36 & 15 & 24 & 23 & 38 \\ 
        9 & 54 & 25 & 22 & 30 & 32 & 30 & 24 \\ 
        10 & 38 & 28 & 26 & 21 & 27 & 33 & 27 \\ \hline
    \end{tabular}
    \vspace*{5mm}
    \caption{Number of correct predictions on the MLQA-dev dataset distributed within the top-10 positions and across different question languages.}
    \label{tab:top-k-answers}
\end{table}

On one hand, these initial findings in the zero-shot scenario validate prior research on cross-lingual QA knowledge acquisition by multilingual models fine-tuned with English data, such as mBERT, which tends to prioritize languages closely related to English over more distant ones. Nevertheless, despite the suboptimal performance in languages other than English, our analysis reveals the presence of valuable cross-lingual knowledge within the top 10 ranked positions, which holds promise for enhancing cross-lingual transfer. In the forthcoming sections, we put together various ideas and methodologies rooted in these observations, with the aim of enhancing generalized cross-lingual QA transfer.
\subsection{Cross-lingual Sampling}
\label{sec:cross-lingual-sampling}
Recalling Section \ref{sec:extactive-qa-task} on the extractive QA task in \ref{sec:extactive-qa-task}, we denote with $x_{i, j}=(q_i, (c_j, a_j))$ an example, where the $i$ indicates the language of the question $q$, and $j$ indicates the language of the context $c$ and the answer $a$, a notation suitable for cross-lingual applications. Here, the first index indicates the language of the question, and the second index indicates the language of the context (and the language from which the answer is extracted). Assuming the availability of a parallel QA dataset, where examples are aligned across multiple, usually via a translation process, we construct semantically equivalent QA examples by solely mixing question and answer languages.

Formally, given a set of seed examples $N_{seed}$ in a source language and their translations into $N_{tl}$ target languages, we randomly sample $ntl$ target languages and generate all possible cross-lingual combinations for each example. By following this approach, the total number of sampled cross-lingual examples grows quadratically with respect to the number of target languages, following the formula:
\begin{equation}
    N_{cross-lingual} = N_{seed} \cdot (1 + {ntl})^2
    \label{data-sampling-eq}
\end{equation}

Our approach leads to a quadratic increase in the number of potentially relevant examples, allowing for cross-lingual fine-tuning with mixed question and context languages across $N_{tl}$ target languages. Consequently, our method encourages generalized cross-lingual performances directly at the level of data. Note that we anticipate a certain degree of redundancy as the number of sampled target languages $ntl$ increases, potentially leading to overfitting. Therefore, in the forthcoming sections, we conduct experiments by varying the $nlt$ variable and analyze its impact on the model performance.

\subsection{Self-knowledge Distillation Objective} \label{KD}
Our work is based on the knowledge distillation techniques \cite{Hinton2015DistillingTK} to perform effective cross-lingual transfer for the extractive QA task. Specifically, we employ self-distillation techniques in generations methods \cite{pmlr-v80-furlanello18a,noauthor_training_nodate,zhang_self-distillation_2020}. Here, both the teacher and student models share identical architectures and sizes, and the teacher's parameters are synchronized with the student's parameters after specific steps or epochs while maintaining a fixed relationship between them.

Following the Equation \ref{prob_start}, we indicate with $\hat{p}_{k}^{start}$ and $\hat{p}_{k}^{end}$ the start and end probability of the teacher model (also referred as soft-labels), and with $a_k^{start}$ and $a_k^{end}$ given the ground-truth labels. We then define a knowledge distillation term expressed by the Kullback-Leibler divergence (KL) between the teacher's soft labels $\{\hat{p}_i^l\}$ and the ground-truth labels $\{a_i^l\}$. Hence, for each QA example, the total loss is a linear combination of the sums over the context's tokens $i \in {1, N}$ for the start and end positions $l \in \{start, end\}$ of the cross-entropy and Kullback-Leibler, as follows in the equation below:
\begin{equation} 
\begin{split}
    L_{skd} =  \alpha_{ce} \sum_{l=start}^{end} \hspace{1mm} \sum_{i=1}^{N} \hspace{1mm} log(p_i^l) +  \hspace{1mm}  \alpha_{ce} \sum_{l=start}^{end} \sum_{i=1}^{N} \hspace{1mm} \hat{p}_i^l \hspace{1mm} log(p_i^l) \\
    = \alpha_{ce} \sum_{l=start}^{end} \hspace{1mm} \sum_{i=1}^{N} \hspace{1mm} CE(a_i^l, p_i^l)  +  \hspace{1mm}  \alpha_{kl} \sum_{l=start}^{end} \sum_{i=1}^{N} \hspace{1mm} KL(\hat{p}_i^l, p_i^l))
\end{split}
\end{equation}
\label{eq:skd-loss}

The $\alpha_{ce}$ and $\alpha_{kl}$ are hyperparameters that represent coefficients to weight the contribution of each loss term independently. Finally, in our self-distillation in generations implementation,  we update the teacher's parameter with the student's parameters after each epoch, as follows:

$$
\begin{cases}
    \{\hat{p}_i^l\} = \{p_i^l\},  & \text{at each epoch}\\
    \{\hat{p}_i^l\} \neq \{p_i^l\},  & \text{between epochs}
\label{data-sampling}
\end{cases}
$$

\subsection{mAP@k Coefficients to Adjust Knowledge Transfer}
\label{mAP@k-coefficient}
\begin{figure}[!ht]
    \centering
     \includegraphics[scale=0.6]{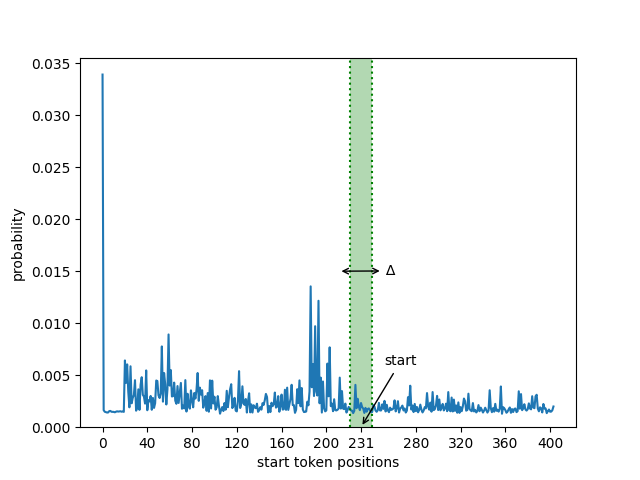}
     \caption{Probability distribution for the start position of the mBERT model on 
     an example with a low F1 score. The figure shows a $\Delta$ interval of several tokens around the start ground truth start. As expected, the peaks of the distribution are not located around the ground truth thus producing a low F1 score.}
     \label{fig:map_at_k_position}
\end{figure}

In the Equation \ref{eq:skd-loss} we formulate the total loss as a linear combination with coefficients $\alpha_{ce}$ and $\alpha_{kl}$, which serve as hyperparameters to weigh the contributions of the KL and CE loss terms. The proper configuration of these coefficients is essential for achieving an optimal learning process. Although various approaches for knowledge distillation have been developed in different deep learning applications \cite{sivasubramanian_alm-kd_2022,wen_preparing_2021}, finding a dynamic method suitable for general cases remains an open problem. 
In this study, we propose a customized technique specifically tailored for the extractive QA task based on the {\em{mAP@k coefficients}}. Our approach dynamically adjusts the loss coefficients for each batch of examples to address the issue of incorrect or informative teacher predictions that may lead to weak student training. We base our technique on the insights gained from the analysis conducted in Section \ref{sec:zero-shot}. Specifically, we observe that the teacher model's degree of cross-lingual QA knowledge is not evenly distributed across different combinations of question-and-answer languages. Additionally, we find that a significant number of correct predictions are spread among the top-10 ranked positions. The uneven and poorly ranked cross-lingual knowledge of the teacher model can result in inaccurate predictions, negatively impacting the performance of the student model. In addressing this challenge, our objective is to create a dynamic formula for selecting teacher-generated predictions that exhibit acceptable cross-lingual knowledge. Subsequently, we enhance these selected predictions by incorporating the cross-entropy term derived from the ground-truth hard labels. We accomplish this task by adopting an information retrieval perspective. Specifically, we introduce a heuristic criterion for identifying the relevance of a teacher's prediction: it is deemed relevant if it falls within a predefined interval around the actual ground truth position. Consequently, we utilize the count of relevant predictions within the top $k$ predictions as a surrogate measure for assessing the quality of the teacher's probability distribution. This assessment, in turn, guides the weighting of the corresponding loss term.
We visualize an example of this interval for the probability distribution on the start position in Figure 
\ref{fig:map_at_k_position}.

More formally, for each QA example, and each start and end position over the context, we calculate the well-established Mean Average Precision at k (mAP@k) metric to define the $\alpha_{kl}$ coefficient, as provided by the equation below:

\begin{equation}
    \alpha_{kl}^{l} \coloneqq
    \frac{\sum_{i=1}^{N} AP_k}{N} =  \frac{\sum_{i=1}^{N} (\frac{\sum_{j=1}^{k} P(j) \hspace{1mm} \delta(j \in s\pm\Delta_s)}{2\Delta_s})}{N}
\label{eq:mapk}
\end{equation}
In the above Equation, $N$ denotes the total number of examples, k is the rank of predictions at which we stop, $P(j)$ represents the precision at the cut-off position $j$ and the indicator function $\delta(j \in l\pm \Delta_l)$ takes the value of 1 if the prediction at position $j$ fall within an interval of length $2\Delta_l$ centred around the ground truth start or end position $l$. By applying these dynamic coefficients to the self-knowledge distillation equation in \ref{eq:skd-loss}, for each QA example, we compute the total loss as follows:
\begin{equation} 
\begin{split}
    L_{skd}^{mAP@k} =\sum_{l=start}^{end} \hspace{1mm} \sum_{i=1}^{N} \hspace{1mm} CE(a_i^l, p_i^l)  +  \hspace{1mm}  \alpha_{kl}^{mAP@k} \sum_{l=start}^{end} \sum_{i=1}^{N} \hspace{1mm} KL(\hat{p}_i^l, p_i^l))
\end{split}
\end{equation}
Where $\alpha_{kl}^{mAP@k}$ is computed as the average of the $\alpha_{kl}^{l}$ and $\alpha_{kl}^{l}$ coefficients, and $\alpha_{ce}$ is set to 1. The relevance of the aforementioned equation lies in the dynamic scaling of the self-knowledge distillation terms, which is directly proportional to the quality of the teacher's predictions. Consequently, higher coefficients result in a more substantial contribution to the overall loss. We assert that employing the mAP@k metric allows for a more accurate assessment of the teacher's high-quality distribution, which is crucial for optimizing cross-lingual transfer efficiency.

\section{Experimental Setting}
We provide details of the experimental setting encompassing the datasets, the training details and the evaluations conducted.

\subsection{Datasets}
For our investigation into cross-lingual transfer in the extractive QA task and the evaluation of our models across diverse languages, we utilized three well-suited datasets.

\paragraph{XQUAD}
We incorporate the XQuAD dataset \cite{artetxe_cross-lingual_2020} consisting of 240 paragraphs and 1,190 question-answer from SQuAD v1.1 \cite{rajpurkar_squad_2016} pairs translated into ten languages: Spanish, German, Greek, Russian, Turkish, Arabic, Vietnamese, Thai, Chinese, and Hindi. These translations were performed by professional human translators. 

\paragraph{MLQA}
The MLQA benchmark \cite{lewis_mlqa_2020} includes QA instances in seven languages: English, Arabic, German, Spanish, Hindi, Vietnamese, and Simplified Chinese. It comprises over 12,000 instances in English and approximately 5,000 instances in each of the other languages, with an average of four language pairs for each instance. The dataset is divided into development and test splits, consisting of 4,199 and 41,244 examples, respectively. The creation of MLQA involved a meticulous process encompassing professional translation and human annotation. It consists of three steps: parallel sentence mining, English QA annotation, and target language QA annotation. MLQA aims to drive research in cross-lingual QA and bridge the gap between training and testing language performance.

\paragraph{TyDiQA-goldp}
TyDiQA \cite{clark_tydi_2020} is a comprehensive question-answering dataset that spans 11 typologically diverse languages and includes three tasks: Passage Selection (SelectP), Minimal Answer Span (MinSpan), and Gold Passage (GoldP). For our study, we focus on the Gold Passage task (GoldP), which involves predicting the contiguous span of characters that answer the question when a passage containing the answer is provided. This task enables comparison with prior works and allows compatibility with existing datasets for extractive QA.  The dataset aims to evaluate the models's ability to generalize across a wide range of languages by including linguistic phenomena that are not typically found in English-only corpora. Additionally, the dataset is collected directly in each language, without relying on translations.

In our work, we use the XQuAD for training, selecting the languages present in MLQA. The MLQA-dev benchmark serves as the development set for model selection and hyperparameter tuning. Finally, we employ the MLQA-test, the remaining portion of XQuAD, and TyDiQA-goldp for evaluation. This selection is based on several considerations:

\begin{itemize}
    \item The XQuAD dataset, with a limited number of samples (1090 examples), is suitable for exploring training with a restricted number of aligned labelled examples across multiple languages. 
    Since it is fully aligned across a wide range of languages, it represents a suitable choice to apply our cross-lingual fine-tuning approach. Additionally, its creation process is straightforward, involving the translation of the original SQuAD dataset without requiring annotation.
    \item The MLQA-test dataset provides an ideal evaluation framework for assessing cross-lingual question-answering performance. It supports both the standard cross-lingual transfer (XLT) task and the more challenging generalized cross-lingual transfer (GXLT) task, which involves different question and context languages. Importantly, to our knowledge, the MLQA-dev set has not been used by the community for cross-lingual QA transfer experiments, even though it is valuable for hyperparameter optimization and model selection as emphasized by the dataset authors.
    \item The TyDiQA-goldp dataset provides an excellent zero-shot testbed for thoroughly evaluating the generalization capabilities of our cross-lingual fine-tuning approach on languages not encountered during training. Importantly, it avoids the use of translation, better reflecting the characteristics of the languages and reducing the potential exploitation of translation artifacts to enhance system performance, as discussed in \cite{artetxe-etal-2020-translation}.
\end{itemize}

\subsection{Training}
The training process is divided into two phases, namely: 1) large-scale QA fine-tuning, involving the use of numerous QA examples to imbue the model with proficient QA capabilities, and 2) cross-lingual QA fine-tuning, designed to enhance and amplify cross-lingual QA transfer by utilizing few labelled QA examples that are aligned across multiple languages. In the first phase, we conduct fine-tuning on the widely adopted multilingual language model, mBERT \cite{devlin_bert_2019}, for the extractive QA task. We solve the task as in Section \ref{sec:extactive-qa-task} and implement the loss functions defined in Equations \ref{eq:skd-loss}, setting the coefficients $\alpha_{ce}$ and $\alpha_{kl}$ to 1. The optimization is performed using stochastic gradient descent with the Adam optimizer \cite{adam}, utilizing a learning rate of $10^{-3}$, a batch size of 12, and a maximum sequence length of 384 tokens. The training is conducted over 3 epochs. 

In the second phase, we continue fine-tuning the previously trained model, named mBERT-qa-en, by leveraging the QA data sampled from the XQuAD dataset with the cross-lingual sampling described in Section \ref{sec:cross-lingual-sampling} obtaining a total number of cross-lingual examples $N_{cross-lingual}$ that follow the Equation \ref{data-sampling-eq}. We maintain the same hyperparameters as in the previous phase and evaluate the model's performance on the MLQA-dev set. We assess the average cross-lingual transfer performance across all languages by computing both the XLT and G-XLT F1 and EM metrics on the concatenation of all cross-lingual examples in the MLQA-dev split. Model selection is then performed by choosing the set of hyperparameters that yield the highest G-XLT F1 accuracy. 

\subsection{Evaluation}
Our evaluation is categorized into two types: in-language, which pertains to languages utilized during cross-lingual fine-tuning, and zero-shot, as the name implies, encompasses languages that were not encountered during the training phase. In the in-language evaluation, we use the MLQA languages for in-language evaluations on the XLT and G-XLT tasks. We recall that the former evaluation involves examples with the same question and context language, while the latter uses examples with different question and context languages. We compute the standard F1 score using the official MLQA evaluation script\footnote{https://github.com/facebookresearch/MLQA/blob/main/mlqa\_evaluation\_v1.py} to properly handle language-specific modifications. In the zero-shot evaluation, we employ the TyDiQA-goldp and the remaining languages of XQuAD datasets that we do not utilize during training. 
\subsubsection{Baselines}
To ensure a direct comparison with our models, we consider as pertinent baselines the methods that involve the mBERT model, applying the same task resolution and evaluating the XLT and G-XLT tasks for the in-language and zero-shot settings. Thus, we exclude approaches that add other variables in terms of training techniques or models. Therefore, to the best of our knowledge, we identify only two baselines that match the previous criteria. The first baseline is our implementation of the mBERT-qa-en used as a zero-shot baseline, as proposed in \cite{lewis_mlqa_2020}, to point out the sheer gain in the performance when cross-lingual fine-tuning is used. Moreover, we adopt as a strong baseline the best-performing mBERT model in \cite{bornea_multilingual_2021}\footnote{We refer to the model obtained with the LAF, PSA+QS (en-all) method, as reported in the original paper.}. These models were tailored for cross-lingual QA through fine-tuning of the mBERT model. Subsequently, they are evaluated on both the MLQA dataset and the TyDiQA-goldp benchmark to assess their performance in terms of XLT and G-XLT tasks, spanning both in-language and zero-shot settings.

\section{Results and Discussions}
Following the setting introduced in the previous section, we compare the best model configurations on the development of the MLQA dataset for the standard cross-entropy and self-distillation training, as described in Section 4. Since we are primarily interested in the generalized cross-lingual QA performance, we consider as best models the ones obtaining the highest F1 score for the G-XLT metric.

\subsection{In-Language Performance}
\begin{figure}[ht!]
     \centering
     \begin{subfigure}[b]{0.49\textwidth}
         \centering
         \includegraphics[width=\textwidth]{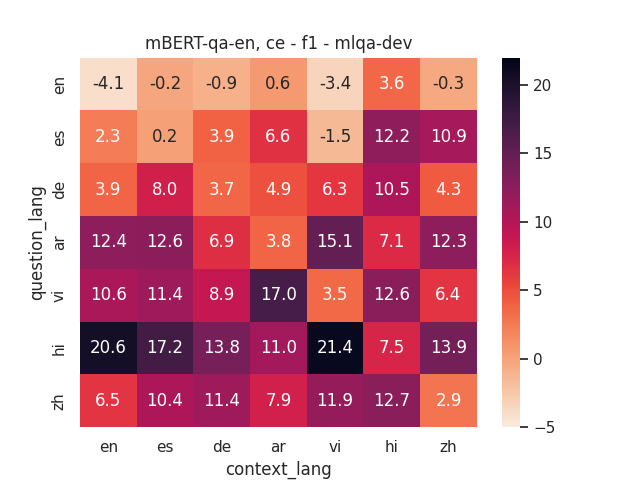}
     \end{subfigure}
     \begin{subfigure}[b]{0.49\textwidth}
         \centering
         \includegraphics[width=\textwidth]{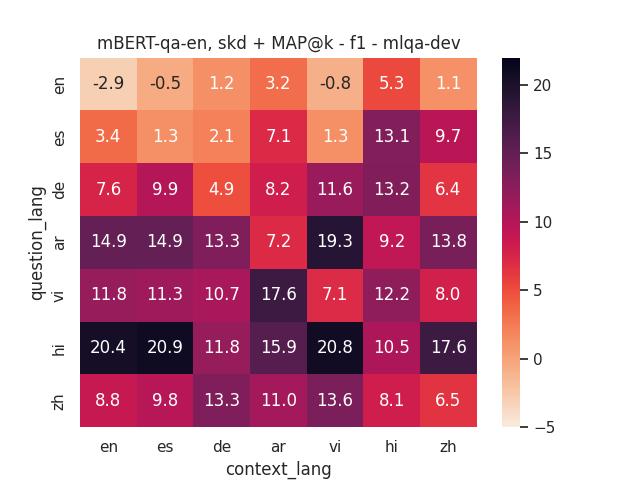}
     \end{subfigure}
     \centering
     \begin{subfigure}[b]{0.49\textwidth}
         \centering
         \includegraphics[width=\textwidth]{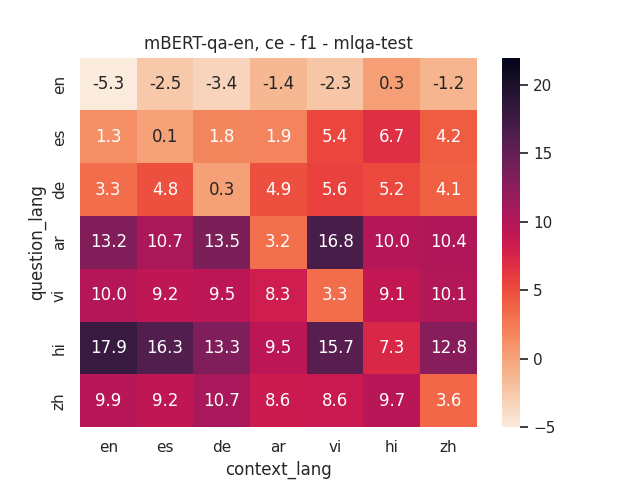}
     \end{subfigure}
     \begin{subfigure}[b]{0.49\textwidth}
         \centering
         \includegraphics[width=\textwidth]{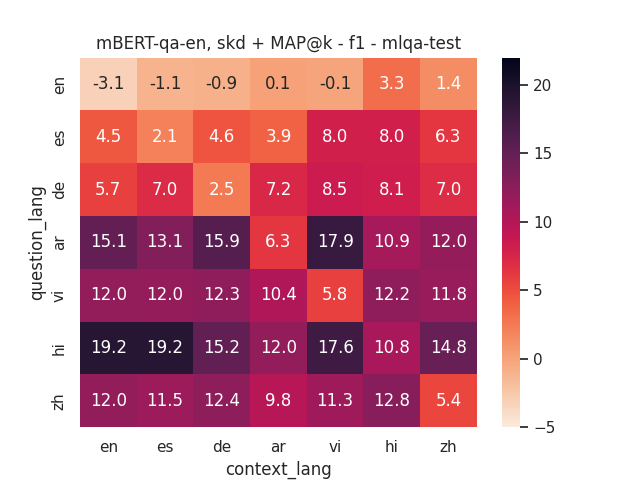}
     \end{subfigure}
        \caption{Difference in F1 scores on the MLQA-dev and MLQA-test sets between the zero-shot performances of the mBERT-qa-en and our best models, namely, a) cross-entropy and b) self-knowledge distillation plus mAP@k.}
        \label{fig:heatmap-mlqa}
\end{figure}

We present the results in various tables and figures, providing both average scores, by taking the average performance across questions in all available languages for each answer language, and single-language scores, using examples with the same question and answer language. Table \ref{table:in-language-global-scores} and \ref{in-language-single-scores} demonstrate the superiority of our distillation methods over the standard cross-entropy fine-tuning approach. The distillation methods exhibit a substantial improvement of more than 2 F1 points in the XLT and G-XLT scores for both the MLQA dev and test sets. Compared to the zero-shot baseline, the distillation methods achieve a remarkable gain of over 9 F1 points in the G-XLT scores for both the dev and test sets of MLQA. Crucially, the results highlight the benefits of utilizing the mAP@k technique to balance the loss terms, leading to improved average cross-lingual scores. The distillation approaches consistently outperform the cross-entropy and zero-shot baselines at the single-language level. Importantly, we are able to achieve more than 90\% F1 score on the MLQA-test set over the strong baseline established in \cite{bornea_multilingual_2021} despite utilizing only 3.5\% of their training data\footnote{The percentage is calculated as the ratio of the $N_{cross-lingual}$ between our methods and the strong baseline, as reported in Table \ref{table:in-language-global-scores}.}. This observation emphasizes the effectiveness of our approaches in scenarios where obtaining large-scale and high-quality translated data is challenging or unfeasible. Additionally, to further investigate the G-XLT capabilities achieved by our method, we examine how the improvement in F1 scores is distributed across different language pairs. Thus, we compare the F1 scores of our best-fine-tuned models with the zero-shot performances of the mBERT-qa-en model for each pair of languages. To visualize these differences, we use heatmaps depicting the variations in F1 scores across language pairs, as shown in Figure \ref{fig:heatmap-mlqa}. The heatmaps clearly demonstrate the superiority of our self-distillation methods over standard cross-entropy fine-tuning, particularly in dissimilar language pairs such as hi-es and ar-vi. Notably, when the question is presented in English, the self-distillation fine-tuning approach exhibits a significantly reduced decline in performance, indicating its propensity to transfer the English QA knowledge to other languages while maintaining its initial performance. In Appendix \ref{fig:heatmap-mlqa-em}, we report an analogous trend for the EM score. This finding provides further evidence to support the higher average G-XLT values discussed earlier, thereby reinforcing the notion of enhanced generalized cross-lingual transfer achieved by our proposed methods.

\begin{table}[!ht]
    \centering
    \resizebox{0.9\textwidth}{!}{\begin{tabular}{l|lll|cc|cc} \hline
    &&&&\multicolumn{2}{c}{MLQA-dev}&\multicolumn{2}{|c}{MLQA-test} \\
        model & ntl & t & $N_{cross-lingual}$ & GXLT & XLT & GXLT & XLT \\ \hline
        mBERT-qa-en    & - & - & -     & 49.7/33.6          & 59.7/42.6          & 49.7/33.8          & 59.4/33.8 \\ 
        \hspace{1mm} w/ skd           & 5 & 2 & 42,840 & 59.1/42.6          & 64.1/46.2          & 58.4/41.6
          &	\textbf{64.0/46.3} \\ 
        \hspace{1mm} w/ skd ,  mAP@k   & 5 & 2 & 42,840 & \textbf{59.2/42.7} & \textbf{64.7/47.1} & \textbf{58.5/41.8} & 63.7/45.9 \\
        \hspace{1mm} w/ ce            & 3 & - & 19,040 & 57.5/40.8          & 62.2/44.8          & 56.3/39.8          & 61.2/44.0 \\ \hline
        % mBERT-qa-en    & - & - & -     & 49.7/33.6          & 59.7/42.6          & 49.7/33.8          & 59.4/33.8 \\ 
        LAF, PSA+QS (en-all) \cite{bornea_multilingual_2021} & -  & - & 1,233,776 & - & -               & $\textbf{61.9}^{\dagger}$/-          & $\textbf{65.7}^{\dagger}$/- \\ \hline
    \end{tabular}}
    \vspace*{5mm}
    \caption{Average scores (F1/EM) of the best-on-dev for the G-XLT and XLT tasks on the MLQA datasets. We also provide the number of target languages $ntl$, the temperature $t$ and the total number of cross-lingual fine-tuning examples $N_{cross-lingual}$ for each model. To distinguish the comparisons between our models and the strong baseline from \cite{bornea_multilingual_2021}, we employ the symbol $\dagger$ to represent the scores of the latter.}
    \label{table:in-language-global-scores}
\end{table}

\begin{table}[!ht]
    \resizebox{\textwidth}{!}{\begin{tabular}{l|ccccccc} \hline
        &\multicolumn{7}{c}{MLQA-dev} \\
        model  & en & es & de & ar & vi & hi & zh \\ \hline
        mBERT-qa-en & 59.3/66.8 & 55.1/50.2 & 49.5/41.0 & 43.3/30.4 & 48.5/37.0 & 44.2/35.5 & 48.1/37.5 \\
         \hspace{1mm} w/ skd  & \textbf{68.6/65.0} & \textbf{65.4/51.2} & 57.7/45.5 & 52.1/34.2 & 58.9/40.7& \textbf{54.9}/44.0 & 56.2/43.0 \\ 
         \hspace{1mm} w/ skd, mAP@k   & 68.4/64.6 & 64.8/50.8 & \textbf{57.7/47.3} & \textbf{53.3/36.2} & \textbf{58.9/42.7} & 54.4/\textbf{45.2} & \textbf{57.1/43.2} \\ 
         \hspace{1mm} w/  ce & 66.7/62.4 & 63.6/50.8 & 56.3/45.5 & 50.7/33.0 & 56.1/41.0 & 53.6/42.0 & 55.3/39.3 \\ \hline

    \end{tabular}}
    \newline
    \vspace*{5mm}
    \newline
    \resizebox{\textwidth}{!}{\begin{tabular}{l|cclcccc}\hline
        &\multicolumn{7}{c}{MLQA-test} \\ 
        model  & en & es & de & ar & vi & hi & zh \\ \hline
        mBERT-qa-en & 58.0/67.4 & 53.1/47.8 & 49.8/45.8 & 44.9/30.4 & 50.3/39.0 & 43.8/31.2 & 47.6/37.0 \\ 
        \hspace{1mm} w/ skd  &  \textbf{67.5/64.4} & 62.3/\textbf{49.2} & 58.5/\textbf{48.6} & 51.9/\textbf{37.0} & 59.8/42.8 & \textbf{53.4/41.9} & 55.7/\textbf{40.5} \\
        \hspace{1mm} w/ skd ,  mAP@k   &  67.3/63.9 & \textbf{62.6}/49.1 & \textbf{58.7}/47.8 &\textbf{52.0}/36.6& \textbf{60.1/43.1} & 53.3/40.8 &\textbf{56.0}/40.0 \\ 
        \hspace{1mm} w/ ce & 65.2/61.5 & 60.0/47.8 & 56.4/45.9 & 49.9/33.5 & 57.9/41.0 & 50.7/38.6 & 53.9/39.3 \\ \hline

        LAF, PSA+QS (en-all) \cite{bornea_multilingual_2021} &$\textbf{74.3}^{\dagger}$/-& $\textbf{66.1}^{\dagger}$/- & $\textbf{61.5}^{\dagger}$/- & $\textbf{54.8}^{\dagger}$/- & $\textbf{64.3}^{\dagger}$/- & $\textbf{54.9}^{\dagger}$/- & $\textbf{57.6}^{\dagger}$/- \\ \hline
    \end{tabular}}
    \vspace*{5mm}
\caption{Single-language scores (F1/EM) of the best-on-dev for the G-XLT and XLT tasks on the MLQA datasets. We drop the number of target languages $ntl$, the temperature $t$ and the total number of training examples $N_{cross-lingual}$ for space reasons. The scores are computed by taking the average performance across questions in all available languages for each answer language. To distinguish the comparisons between our models and the strong baseline from \cite{bornea_multilingual_2021}, we employ the symbol $\dagger$ to represent the scores of the latter.}
\label{in-language-single-scores}
\end{table}

\subsection{Zero-shot Performance}
The findings of the zero-shot evaluation, detailed in Table \ref{table:zero-shot-xquad-tydiqa}, underscore the superior efficacy of self-distillation fine-tuning in contrast to standard cross-entropy fine-tuning. This superiority is manifest in the higher XLT F1 scores obtained for both the XQuAD and TyDiQA-goldp benchmarks. This encompasses both the average XLT score and the individual language levels, with a notable margin. Surprisingly, on the TyDiQA dataset, our results also stand in competitive alignment with the robust baseline proposed in \cite{bornea_multilingual_2021}. Specifically, we surpass this baseline by approximately 2-4 F1 points for all languages except Russian (ru) and Swahili (sw), despite having one order of magnitude less training data, as discussed in the preceding section.
Overall, while the substantial performance boost on the XQuAD dataset may be partly due to subtle translation artifacts between training and test sets, as discussed in \cite{artetxe-etal-2020-translation}, our positive results on the TyDiQA-goldp dataset validate the efficacy of our self-distillation approach. Notably, TyDiQA stands apart from MLQA and XQuAD, as it necessitates genuine information-seeking questions crafted by individuals who do not possess the answers. Furthermore, this dataset is sourced directly in each language, obviating the need for translations.

\begin{table}[!ht]
\footnotesize
    \centering
    \resizebox{0.8\textwidth}{!}{\begin{tabular}{l|ccccc}\hline
        &\multicolumn{5}{c}{XQuAD} \\ 
        model & XLT & el & ru & tr & th \\ \hline
        mBERT-qa-en & 55.3/40.9 & 62.7/45.5 & 70.5/53.2 & 51.2/36.9 & 36.6/27.9 \\
         \hspace{1mm} w/ skd & 72.1/60.1 &	\textbf{81.6/67.5} &	86.8/74.9 &	\textbf{71.0/57.0}	 & 48.9/40.9 \\ 
        \hspace{1mm} w/ skd ,  mAP@k & \textbf{72.7/60.8} & 81.4/66.7 &	\textbf{87.6/76.3} 	& 70.5/56.0 & \textbf{51.2/44.1} \\ 
         \hspace{1mm} w/ ce & 68.8/56.2 & 75.1/59.4 & 85.2/72.9 & 66.2/51.8 & 48.7/40.8 \\ \hline
        % mBERT-qa-en & 55.3/40.9 & 62.7/45.5 & 70.5/53.2 & 51.2/36.9 & 36.6/27.9 \\ \hline
    \end{tabular}}
    \newline
    \vspace*{5mm}
    \newline
    \centering
    \resizebox{0.8\textwidth}{!}{\begin{tabular}{l|cccccccc} \hline
        &\multicolumn{8}{c}{TyDiQA-goldp} \\ 
        model & XLT & bn & fi & in & ko & ru & sw & te \\ \hline
        mBERT-qa-en & 54.0 & 55.9 & 54.7 & 58.1 & 47.5 & 64.3 & 50.2 & 47.3 \\
         \hspace{1mm} w/ skd & \textbf{59.5} & 60.9 & \textbf{61.6} & 65.4 & 55.1 & 65.0 & 59.0 & 49.3 \\ 
         \hspace{1mm} w/ skd ,  mAP@k & 58.9 & \textbf{63.3} & 54.8 & \textbf{68.7} & 53.8 & \textbf{65.0} & 55.3 &\textbf{51.0} \\ 
         \hspace{1mm} w/ ce & 58.6 & 58.1 & 59.0 & 66.4 & \textbf{55.8} & 64.1 & 56.7 & 50.0 \\ \hline
        % mBERT-qa-en & 54.0 & 55.9 & 54.7 & 58.1 & 47.5 & 64.3 & 50.2 & 47.3 \\ 
        LAF, PSA+QS (en-all) \cite{bornea_multilingual_2021} & $\textbf{61.9}^{\dagger}$ & 59.9 & 57.3 & 64.1 & 55.3 & $\textbf{65.9}^{\dagger}$ & $\textbf{63.8}^{\dagger}$ & 49.1 \\ \hline

    \end{tabular}}
    \vspace*{5mm}
    \caption{Single-language zero-shot F1/EM scores of the best-on-dev for the XQuAD and F1 score on the TyDiQA-goldp datasets. To distinguish the comparisons between our models and the strong baseline from \cite{bornea_multilingual_2021}, we employ the symbol $\dagger$ to represent the scores of the latter.}
    \label{table:zero-shot-xquad-tydiqa}
\end{table}
\newpage
\section{Analysis}
To understand the relationships and impact on the training performance of crucial variables such as the number of parallel target languages $ntl$ in cross-lingual sampling, the temperature $t$ in knowledge distillation, and the mAP@k loss coefficients $\alpha_{kl}^{mAP@k}$, we utilise the experiments on the development set of the MLQA dataset and perform analysis on them.

\subsection{Evolution of the mAP@k Coefficients}
\label{sec:map-at-k-evolution}
\begin{figure}[h!]
     \centering
     \includegraphics[width=0.60\textwidth]{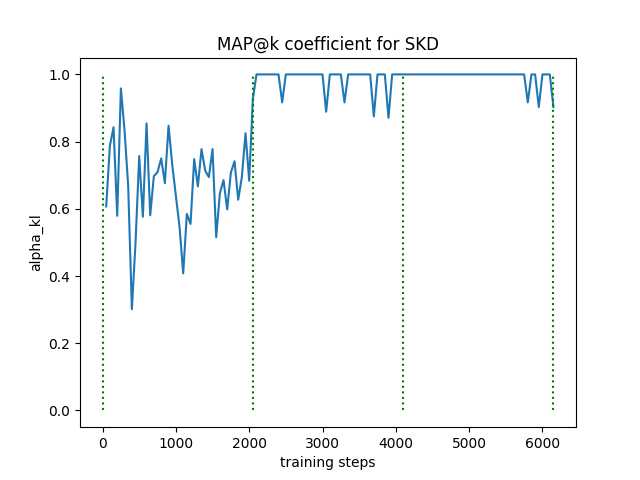}
     \caption{Evolution of the $\alpha_{kl}^{mAP@k}$ coefficient during training. Each vertical line indicates a training epoch.}
     \label{fig:curve-mAP@k-alphas-alpha_kl}
\end{figure}
We commence by investigating the temporal evolution of the mAP@k coefficient, denoted as $\alpha_{kl}^{mAP@k}$, throughout the training process to gain insights into its influence on self-distillation fine-tuning. The mAP@k method we propose in Section \ref{mAP@k-coefficient}, dynamically weighs the significance of the teacher's predictions. It is important to emphasize that in our self-knowledge distillation experiments, the teacher is synchronized with the student model at each epoch. Consequently, we anticipate that, given proper learning of the student model, the $\alpha_{kl}^{mAP@k}$ increment its value, thus increasingly relying on improved teacher predictions. Hence, Figure \ref{fig:curve-mAP@k-alphas-alpha_kl} illustrates the gradual augmentation of the coefficient during training, starting from an initial value of approximately 0.6 and progressing until it attains its maximal value of 1 at approximately half of the training duration. Interestingly, we observe oscillatory behaviour in the values of $\alpha_{kl}^{mAP@k}$ during the initial epoch, which subsequently stabilizes over time, as a consequence of successful student training. This empirical evidence is consistent with the overall superior performance achieved by the models employing the mAP@k method, underscoring its efficacy in promoting the regularization of self-knowledge distillation training.

\subsection{Impact of the Number of Parallel Target Languages}
\begin{figure}[h!]
     \centering
     \begin{subfigure}[b]{0.49\textwidth}
         \centering
         \includegraphics[width=\textwidth]{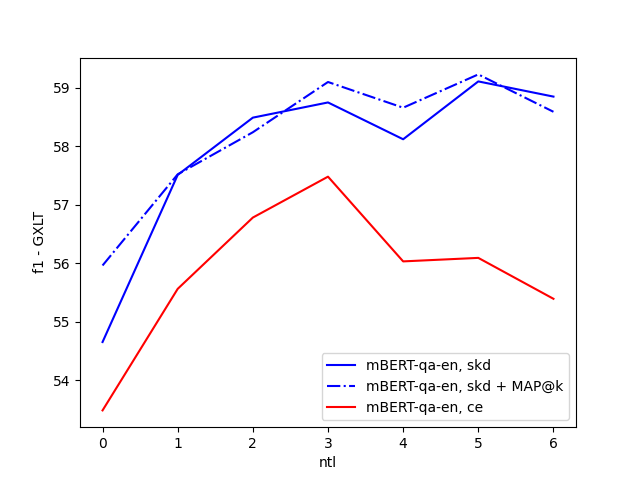}
     \end{subfigure}
     \begin{subfigure}[b]{0.49\textwidth}
         \centering
         \includegraphics[width=\textwidth]{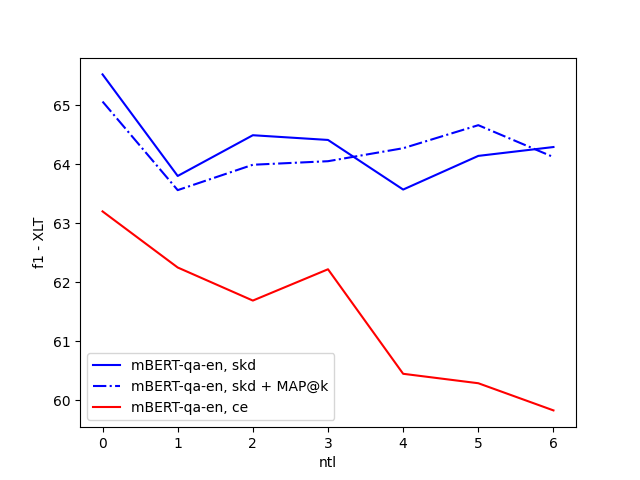}
     \end{subfigure}
        \caption{F1 scores on the MLQA-dev set for the G-XLT and XLT settings as a function of the number of parallel target languages $ntl$ sampled during cross-lingual fine-tuning for our self-distillation models and standard cross-entropy fine-tuning.}
        \label{fig:f1-vs-ntl-mlqa-dev}
\end{figure}
Equation \ref{data-sampling-eq} illustrates how the total number of training examples used for cross-lingual fine-tuning scales quadratically with the number of sampled parallel target languages $ntl$, resulting in a substantial increase in dataset size. Nonetheless, while each combination is, by definition, a distinct QA instance\footnote{effectively, distinct question-context mappings}, we recall that they are all derived from the same set of $N_{seed}$ seed examples in English. Consequently, it is plausible to argue that a multilingual model equipped with powerful cross-lingual representations may map two parallel examples into highly similar representations, thereby not fully harnessing the benefits of these additional examples. Therefore, we aim to assess the impact of $ntl$ in Equation \ref{data-sampling-eq} on the performances of our self-knowledge distillation models. In Figure \ref{fig:f1-vs-ntl-mlqa-dev}, we present the F1 scores of the G-XLT and XLT metrics for both the self-distillation and cross-entropy models, plotted as a function of the number $ntl$ while keeping the temperature set to the default distillation value of 2. The curves corresponding to the G-XLT setting exhibit an increasing trend in F1 scores for the self-distillation models, whereas the cross-entropy models experience a decline for values of $ntl$ greater than 3. Conversely, the XLT metric shows a decreasing trend as the number of parallel target languages increases, with a steep decline observed for the cross-entropy method. Additionally, the curves demonstrate that the self-distillation models using the mAP@k loss coefficients $\alpha_{kl}^{mAP@k}$ achieve better performance for both the G-XLT and XLT scores. We report the corresponding analysis for the EM score in Appendix \ref{app:em-scores}, confirming the same trend. Overall, these results lead us to two conclusions: i) the cross-entropy method tends to overfit when fine-tuning with a cross-lingual sampled dataset, likely due to the increased redundancy of the data; and ii) the self-distillation methods effectively leverage the cross-lingual sampled dataset to enhance G-XLT performance while mitigating overfitting for the XLT metrics, in contrast to the observed trend with the cross-entropy approach.

\subsection{Effect of Temperature on the Number of Target Languages}
\begin{figure}[ht!]
     \centering
     \begin{subfigure}[b]{0.49\textwidth}
         \centering
         \includegraphics[width=\textwidth]{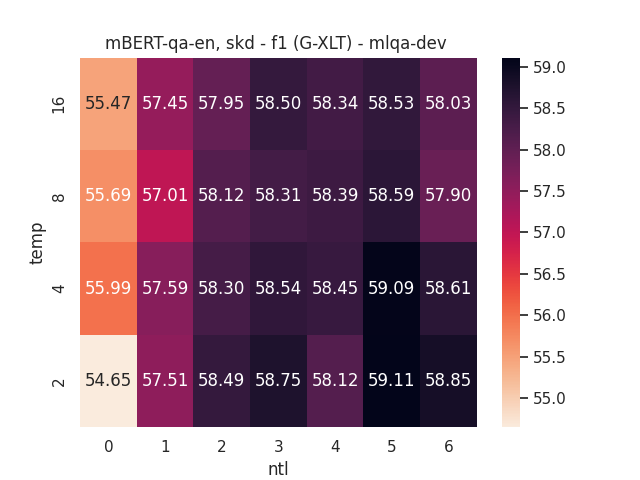}
     \end{subfigure}
     \begin{subfigure}[b]{0.49\textwidth}
         \centering
         \includegraphics[width=\textwidth]{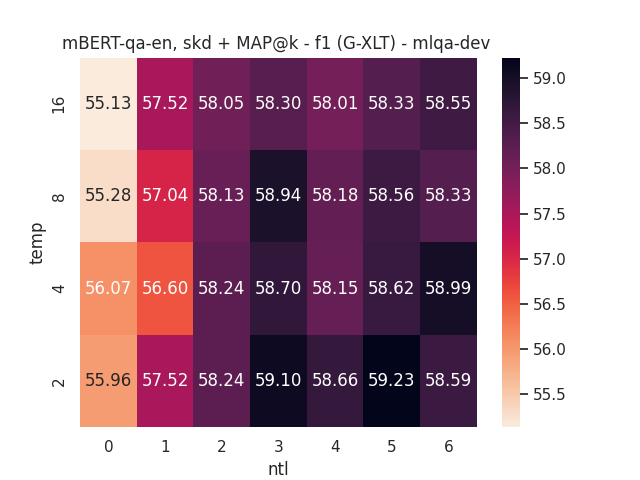}
     \end{subfigure}
        \caption{F1 scores for the G-XLT setting of the self-distillation models on the MLQA-dev set as a function of both the self-distillation temperature and the number of target languages sampled for training. Minimum and maximum values are relative to each figure to better visualise the variation of the scores.}
        \label{fig:heatmap-temp-ntl-mlqa-dev}
\end{figure}

The knowledge distillation temperature $t$ is a key hyperparameter that directly affects cross-lingual transfer by influencing the shape of the teacher's distribution and consequently the Kullback-Leibler term in the loss. In order to explore this impact, we consider the models trained with self-knowledge distillation and investigate the relationship between the F1 score in the G-XLT setting and the number of target languages used in cross-lingual sampling. Figure \ref{fig:heatmap-temp-ntl-mlqa-dev} presents the results of this analysis. We observe that the self-knowledge distillation models improve as the number of target languages $ntl$, observing a sharp increase when $ntl > 2$ while exhibiting relatively small variations as the number of target languages $ntl$ increases. Moreover, the optimal configuration depends on the use of the mAP@k loss coefficients, making it more challenging to determine. We display similar patterns for the EM scores in Appendix \ref{app:em-scores}. Therefore, this analysis highlights the importance of using a number of target languages greater than 2 while underscoring the need for careful tuning of the knowledge distillation temperature to maximize cross-lingual transfer performance.

\section{Ablation Study: Unsupervised Self-Distillation}

\begin{figure}[ht!]
     \centering
     \includegraphics[width=0.6\textwidth]{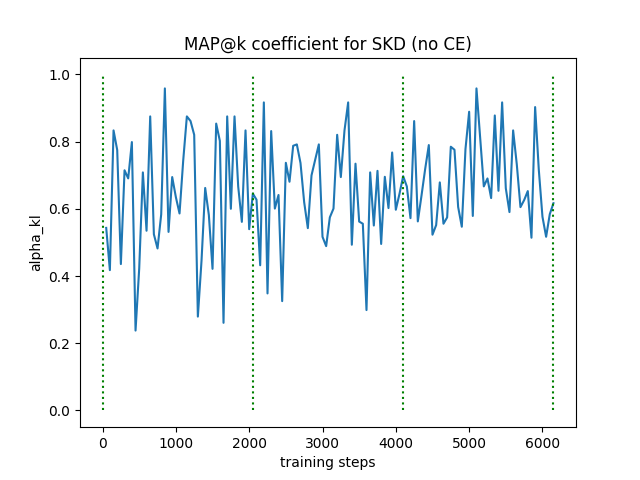}
    \caption{Evolution of the $\alpha_{kl}^{mAP@k}$ coefficient during training. Each vertical line indicates a training epoch.}
    \label{fig:curve-mAP@k-alphas-alpha_kl-no-ce}
\end{figure}
In this section, we investigate the impact of unsupervised self-distillation, where only the Kullback-Leibler divergence term is utilized, thus excluding the cross-entropy term. We conduct an ablation experiment by using our best-performing models in Table \ref{table:in-language-global-scores} and removing the cross-entropy term, The aim is to explore the potential of this unsupervised approach and gain further insights into how to accomplish beneficial cross-lingual fine-tuning using limited labelled examples. The results of the ablation study, presented in Table \ref{table:ablation-study}, reveal a significant decrease in performance by several F1 points in both the in-language and zero-shot evaluations. Surprisingly, the performance deteriorates by several F1/EM points even beyond the baseline achieved by the cross-entropy-only experiments. Furthermore, the inclusion of the mAP@k loss coefficients does not mitigate this performance decline. This observation is supported by the behaviour of the $\alpha_{kl}$ mAP@k coefficient that keeps oscillating around 0.6 and does not exhibit an increasing trend throughout the training process, as depicted in Figure \ref{fig:curve-mAP@k-alphas-alpha_kl-no-ce}. This is in stark contrast to the trend exhibited by the $\alpha_{kl}$ coefficient in Figure \ref{fig:curve-mAP@k-alphas-alpha_kl}, where a progressive increase is evident over the course of training. These findings underscore the critical role played by the cross-entropy term in guiding and enhancing the model's self-knowledge, leading to overall improved cross-lingual transfer performance. We conclude that the inclusion of cross-entropy is crucial for effective knowledge distillation and achieving superior cross-lingual transfer capabilities.

\begin{table}[ht!]
\footnotesize
    \centering
    \resizebox{\textwidth}{!}{\begin{tabular}{l|cccc|cc} \hline
     & \multicolumn{4}{|c}{In-language} &  \multicolumn{2}{|c}{Zero-shot}  \\ 
     & \multicolumn{2}{|c}{MLQA-dev} & \multicolumn{2}{c|}{MLQA-test} & XQuAD & TyDiQA-goldp \\
     model &   GXLT & XLT & GXLT & XLT & XLT & XLT \\ \hline
     mBERT-qa-en & - & - & - & - & - & - \\
     \hspace{1mm} w/ skd,  mAP@k w/o  ce  & 52.5/36.6 & 61.0/43.9
  & 53.1/36.0 & 60.4/43.6  & 57.7/43.6  & 53.9/41.9  \\
     \hspace{1mm} w/ skd w/o  ce  & 53.8/37.9 & 61.2/44.2  & 52.3/37.0 & 60.5/43.6  & 56.8/42.7  & 53.9/41.5 \\
    \hspace{1mm} w/ ce  & \textbf{57.5/40.8} & \textbf{62.2/44.8} & \textbf{56.3/39.8} & \textbf{61.2/44.0} & \textbf{68.8/56.2} & \textbf{58.6/45.3} \\ \hline
    \end{tabular}}
    \vspace*{5mm}
    \caption{Average scores (F1/EM) of the best-on-dev models without the cross-entropy term for the MLQA dev, MLQA test, XQuAD and TyDiQA-goldp datasets. We intentionally do not report the mBERT-qa-en scores to better stress the comparison between the models in the ablation study.}
    \label{table:ablation-study}

\end{table}

\section{Limitations and Future Works}
Our study on cross-lingual question-answering (QA) techniques has shown encouraging results, nevertheless, we acknowledge the main limitations:

\paragraph{Availability of Parallel QA Datasets}
Although our experiments employed as few as 1 thousand examples per target language, the availability of high-quality parallel QA datasets ultimately relies on human annotations that could be resource-intensive for a wide range of languages. Therefore, The scarcity of such data may limit the generalizability and scalability of our approach to a larger set of low-resource languages.

\paragraph{Challenges with Distant Language Families}
Our methods assume that the underlying structure and patterns of the source language, English in our case, can be effectively transferred to target languages. However, as expected, our findings indicate that achieving high-quality cross-lingual QA transfer becomes more challenging when dealing with languages from distant language families. We believe the linguistic dissimilarities among these languages can hinder effective knowledge transfer, leading to performance degradation in specific language pairs.

\paragraph{Sensitivity to the Hyperparameters}
Our self-distillation techniques and cross-lingual sampling performance are sensitive to various method-specific training hyperparameters. Determining the optimal values for these hyperparameters requires proper tuning.

Acknowledging these limitations and addressing them through future research will contribute to further advancements in cross-lingual QA, allowing for more effective and efficient cross-lingual QA across a diverse and low-resource range of languages.
\section{Conclusion}
This study presents an effective approach for achieving cross-lingual QA transfer abilities with limited data aligned across languages. By leveraging advanced self-knowledge distillation techniques and cross-lingual sampling, our method offers promising prospects for bridging the language gap in QA tasks and enhancing G-XLT performances in data-scarce scenarios. Importantly, the introduction of a novel dynamic loss weighting technique with mAP@k coefficients adds an essential contribution to cross-lingual transfer methods, allowing for more optimized knowledge distillation during fine-tuning. Therefore, our approach proves beneficial for languages where high-quality machine-translated data is scarce, but small-scale annotation efforts are feasible. Overall, our research contributes to the advancement of cross-lingual QA models, providing valuable insights and methodologies for future work in this domain.

\section*{Acknowledgments}
This work was supported by the project PID2019-107579RB-I00 (MICINN).

%Bibliography
\bibliographystyle{unsrt}  
\bibliography{references}  

\newpage
\appendix
\section*{Appendix A: Reproducibility and Accessibility}

To guarantee reproducibility, we open-source under the Apache License 2.0 the code used to run the experiments: \href{https://github.com/ccasimiro88/self-distillation-gxlt-qa}{GitHub repository}. Moreover, we release our best self-knowledge distillation models under the Apache License 2.0 and upload them on the HuggingFace Model Hub: \href{https://huggingface.co/ccasimiro/mbert-qa-en-skd-self-distill}{mBERT-qa-en-skd} and
\href{https://huggingface.co/ccasimiro/mbert-qa-en-skd-map-coeff-self-distill}{mBERT-qa-en-skd-mAP@k}

\section*{Appendix B: Report on the Exact Match Perfomances}
\label{app:em-scores}
We report the corresponding EM scores relative to the models and experiments of the paper.
\subsection*{Zero-shot Performance}

\begin{figure}[!h]
    \centering
    \begin{subfigure}[b]{0.49\textwidth}
    \includegraphics[scale=0.50]{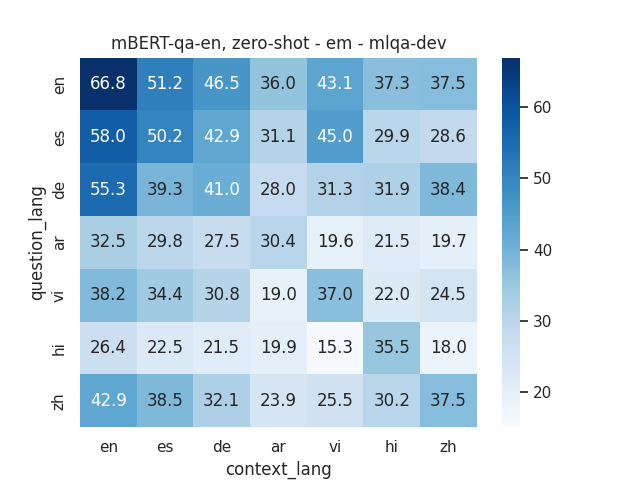}
    \label{fig:mbert-qa-en-zero-shot-baseline}
    \end{subfigure}
    \begin{subfigure}[b]{0.49\textwidth}
    \includegraphics[scale=0.50]{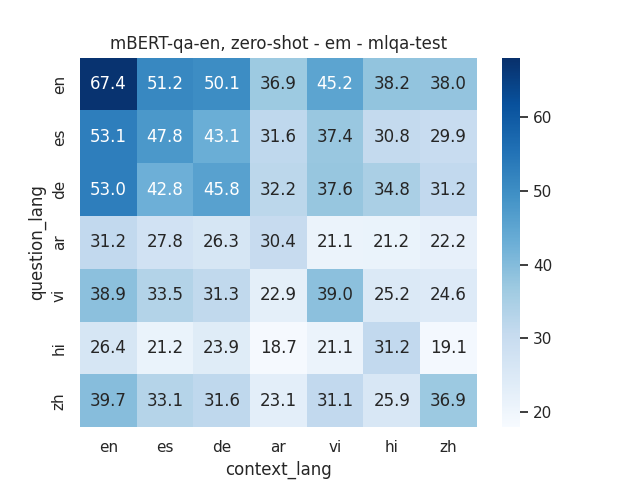}
    \label{fig:mbert-qa-en-zero-shot-baseline}
    \end{subfigure}
    \caption{Zero-shot EM scores on the MLQA-dev and MLQA-test datasets of the mBERT-qa-en model.}
\label{fig:mbert-qa-en-zero-shot-baseline-em}
\end{figure}

Figure \ref{fig:mbert-qa-en-zero-shot-baseline-em} reports the zero-shot EM scores on the MLQA-dev and MLQA-test datasets of the mBERT-qa-en model. We observe how the performance degrades depending on the pair of non-English languages used to test the model, similar to the F1 scores in Figure \ref{fig:heatmap-mlqa}.
\subsection*{Cross-lingual Fine-tuning}
\subsubsection*{Improvements Across Languages}
\begin{figure}[h!]
     \centering
     \begin{subfigure}[b]{0.49\textwidth}
         \centering
         \includegraphics[width=\textwidth]{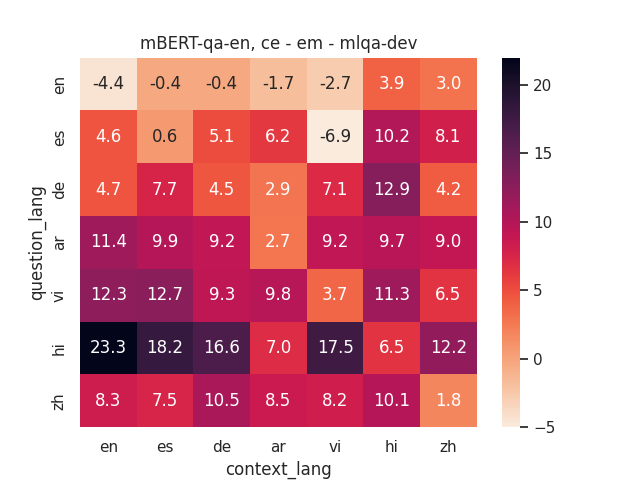}
     \end{subfigure}
     \begin{subfigure}[b]{0.49\textwidth}
         \centering
         \includegraphics[width=\textwidth]{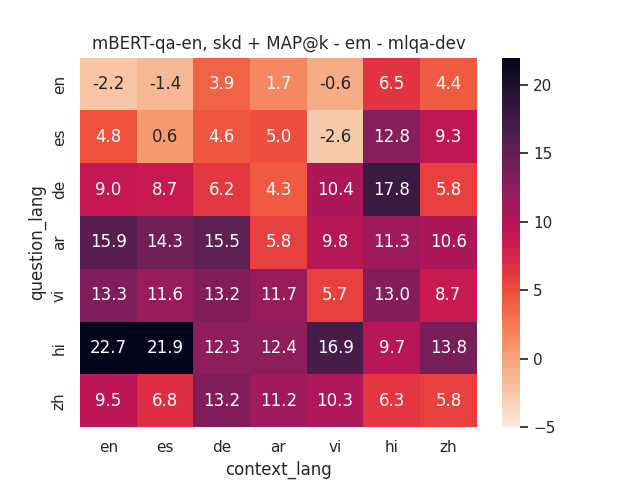}
     \end{subfigure}
     \centering
     \begin{subfigure}[b]{0.49\textwidth}
         \centering
         \includegraphics[width=\textwidth]{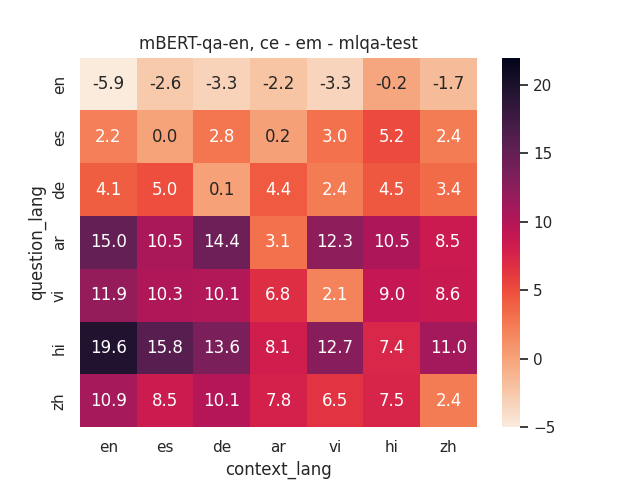}
     \end{subfigure}
     \begin{subfigure}[b]{0.49\textwidth}
         \centering
         \includegraphics[width=\textwidth]{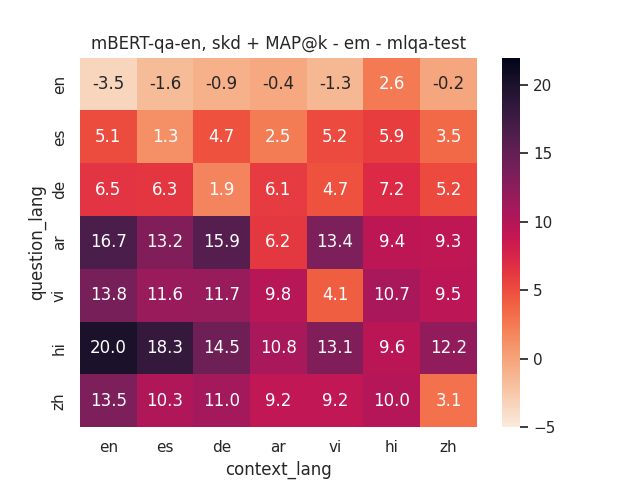}
     \end{subfigure}

        \caption{Difference in EM scores on the MLQA-dev and MLQA-test sets between the zero-shot performances of the mBERT-qa-en and our best models, namely, a) cross-entropy and b) self-knowledge distillation plus mAP@k.}
        \label{fig:heatmap-mlqa-em}
\end{figure}

Figure \ref{fig:heatmap-mlqa-em} shows the difference in EM scores on the MLQA-dev and MLQA-test sets between the mBERT-qa-en zero-shot performances and our best models, namely, a) cross-entropy and b) self-knowledge distillation plus mAP@k. We observe similar patterns as for the F1 scores in Figure \ref{fig:heatmap-mlqa}.

\subsubsection*{Temperature vs. Number Target Languages}
\begin{figure}[!h]
     \centering
     \begin{subfigure}[b]{0.49\textwidth}
         \centering
         \includegraphics[width=\textwidth]{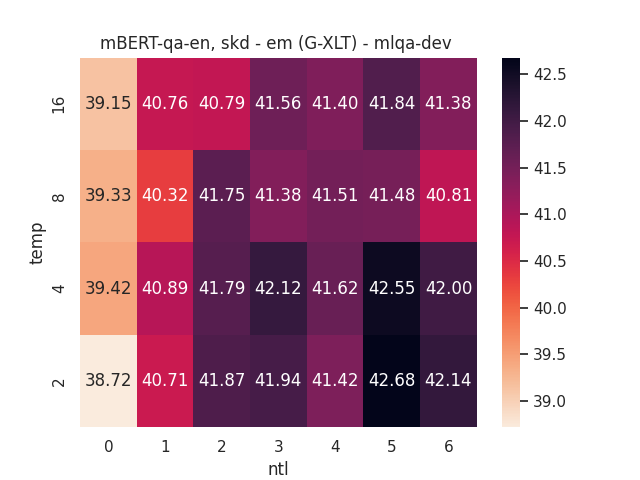}
     \end{subfigure}
     \begin{subfigure}[b]{0.49\textwidth}
         \centering
         \includegraphics[width=\textwidth]{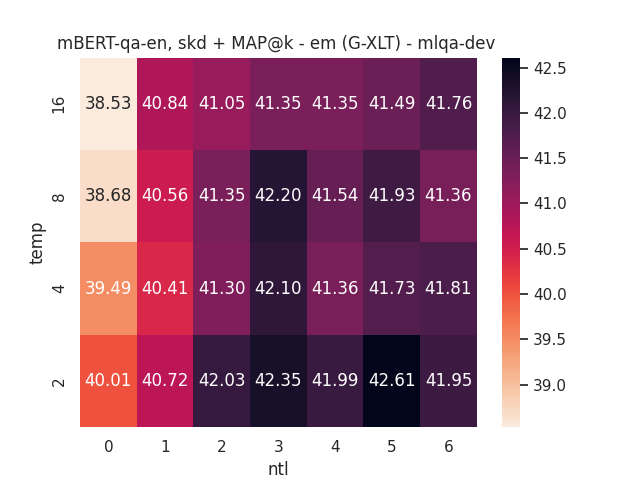}
     \end{subfigure}
        \caption{EM scores for the G-XLT setting of the self-distillation models on the MLQA-dev set as a function of both the self-distillation temperature and the number of target languages sampled for training. Minimum and maximum values are relative to each figure to better visualise the variation of the scores.}
        \label{fig:heatmap-temp-ntl-mlqa-dev-em}
\end{figure}

Figure \ref{fig:heatmap-temp-ntl-mlqa-dev-em} shows F1 scores for the G-XLT setting of the self-distillation models on the MLQA-dev set as a function of both the self-distillation temperature and the number of target languages sampled for training. Also, in this case, we report trends similar to the F1 scores in Figure \ref{fig:heatmap-temp-ntl-mlqa-dev}.

\subsubsection*{Number Target Languages}
\begin{figure}[!ht]
     \centering
     \begin{subfigure}[b]{0.49\textwidth}
         \centering
         \includegraphics[width=\textwidth]{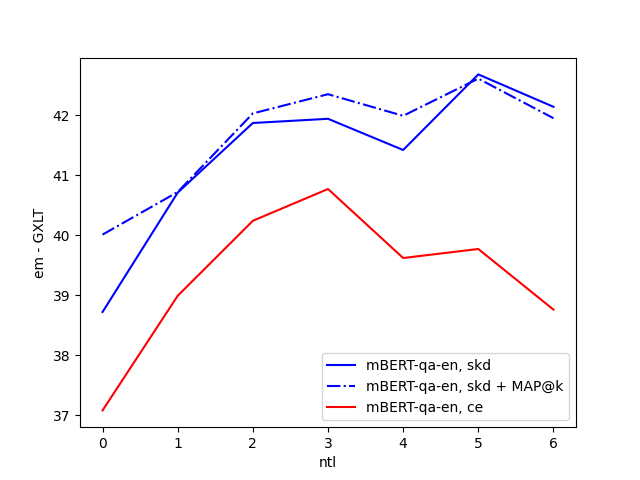}
     \end{subfigure}
     \begin{subfigure}[b]{0.49\textwidth}
         \centering
         \includegraphics[width=\textwidth]{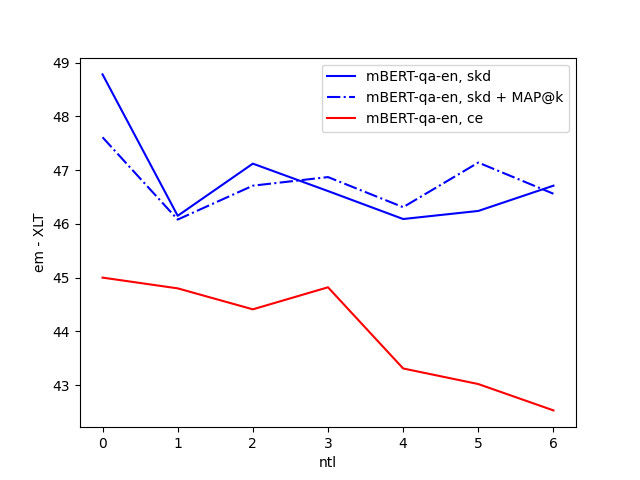}
     \end{subfigure}
     
        \caption{EM scores on the MLQA-dev set for the G-XLT and XLT settings as a function of the number of target languages.}
        \label{fig:f1-vs-ntl-mlqa-dev-em}
\end{figure}

Figure \ref{fig:f1-vs-ntl-mlqa-dev-em} shows the EM scores on the MLQA-dev set for the G-XLT and XLT settings as a function of the number of target languages sampled during cross-lingual fine-tuning for our self-distillation and standard cross-entropy fine-tuning. The EM scores follow a similar pattern as the F1 scores in Figure \ref{fig:f1-vs-ntl-mlqa-dev}.

\end{document}